\documentclass[lettersize,journal]{IEEEtran}
\usepackage{amsmath,amsfonts}
\usepackage{algorithmic}
\usepackage{algorithm}
\usepackage{array}
\usepackage[caption=false,font=normalsize,labelfont=sf,textfont=sf]{subfig}
\usepackage{textcomp}
\usepackage{stfloats}
\usepackage{url}
\usepackage{verbatim}
\usepackage{graphicx}
\usepackage{cite}
\usepackage{bbm}
\usepackage{multirow}
\usepackage{color}
\usepackage{xpatch}
\makeatother

\begin{document}

\title{On the Robustness of Object Detection Models \\ on Aerial Images}

\author{Haodong He, Jian Ding, Bowen Xu, and Gui-Song Xia,~\IEEEmembership{Senior Member,~IEEE}
\thanks{Haodong He, Jian Ding, Bowen Xu, and Gui-Song Xia are with the School of Computer Science, Wuhan University, Wuhan 430072, China (e-mail: haodonghe@whu.edu.cn; jian.ding@whu.edu.cn; xbw@whu.edu.cn; guisong.xia@whu.edu.cn).}% <-this % stops a space
\thanks{Corresponding author: Gui-Song Xia (guisong.xia@whu.edu.cn).}}

\markboth{}
{Shell}

\maketitle

\begin{abstract}
The robustness of object detection models is a major concern when applied to real-world scenarios. The performance of most models tends to degrade when confronted with images affected by corruptions, since they are usually trained and evaluated on clean datasets.
While numerous studies have explored the robustness of object detection models on natural images, there is a paucity of research focused on models applied to aerial images, which feature complex backgrounds, substantial variations in scales, and orientations of objects.
This paper addresses the challenge of assessing the robustness of object detection models on aerial images, with a specific emphasis on scenarios where images are affected by clouds.
In this study, we introduce two novel benchmarks based on DOTA-v1.0. The first benchmark encompasses 19 prevalent corruptions, while the second focuses on the cloud-corrupted condition—a phenomenon uncommon in natural images yet frequent in aerial photography. We systematically evaluate the robustness of mainstream object detection models and perform necessary ablation experiments. Through our investigations, we find that rotation-invariant modeling and enhanced backbone architectures can improve the robustness of models. Furthermore, increasing the capacity of Transformer-based backbones can strengthen their robustness. The benchmarks we propose and our comprehensive experimental analyses can facilitate research on robust object detection on aerial images. The codes and datasets are available at: https://github.com/hehaodong530/DOTA-C.
\end{abstract}


\section{Introduction}
\IEEEPARstart{T}{he} robustness of object detection models on aerial images is crucial for ensuring accurate and reliable results in real-world scenarios. In recent years, models based on deep neural networks have reached the state-of-the-art level~\cite{orientedrcnn,roitrans,redet,sfrnet}, constantly refreshing the highest evaluation scores on open competitions. Despite the high performance on existing aerial object detection datasets, the reliability and applicability of those methods in actual usage remain to be examined.

Widely used datasets for object detection on aerial images such as UCAS-AOD \cite{ucasaod}, HRSC2016 \cite{hrsc}, and DOTA \cite{dota} are all carefully curated and post-processed. They only contain clean and high-quality aerial images. Consequently, most of the current research conducted experiments by training and testing models on the conditions of independent and identically distributed (IID) data. However, in practical scenarios, the 
quality of aerial images is affected by multiple factors including weather conditions, the camera itself, and many others. A taken aerial image is likely to have unknown quality, which means that out-of-distribution (OOD) data are almost always encountered.

Notably, deep learning-based models excelling on IID data often manifest massive performance degradation when confronted with OOD data for visual tasks~\cite{corruptions1,corruptions4}. To assess the robustness of learning-based models for OOD data, several benchmarks have been introduced. For example, the ImageNet-C dataset~\cite{imagenetc} defines 19 common visual corruptions and applies them to ImageNet challenge~\cite{imagenet} to benchmark the robustness of classification models to image corruptions. Similar corruptions are also extended to frequently used object detection datasets like COCO~\cite{lin2014microsoft} as the measurement of general object detection~\cite{autonomousdriving}. They made detailed explorations of natural images in the case of degraded quality and illustrated the importance of robustness for learning-based models. 

\begin{figure}
    \centering
    \includegraphics[width=\linewidth]{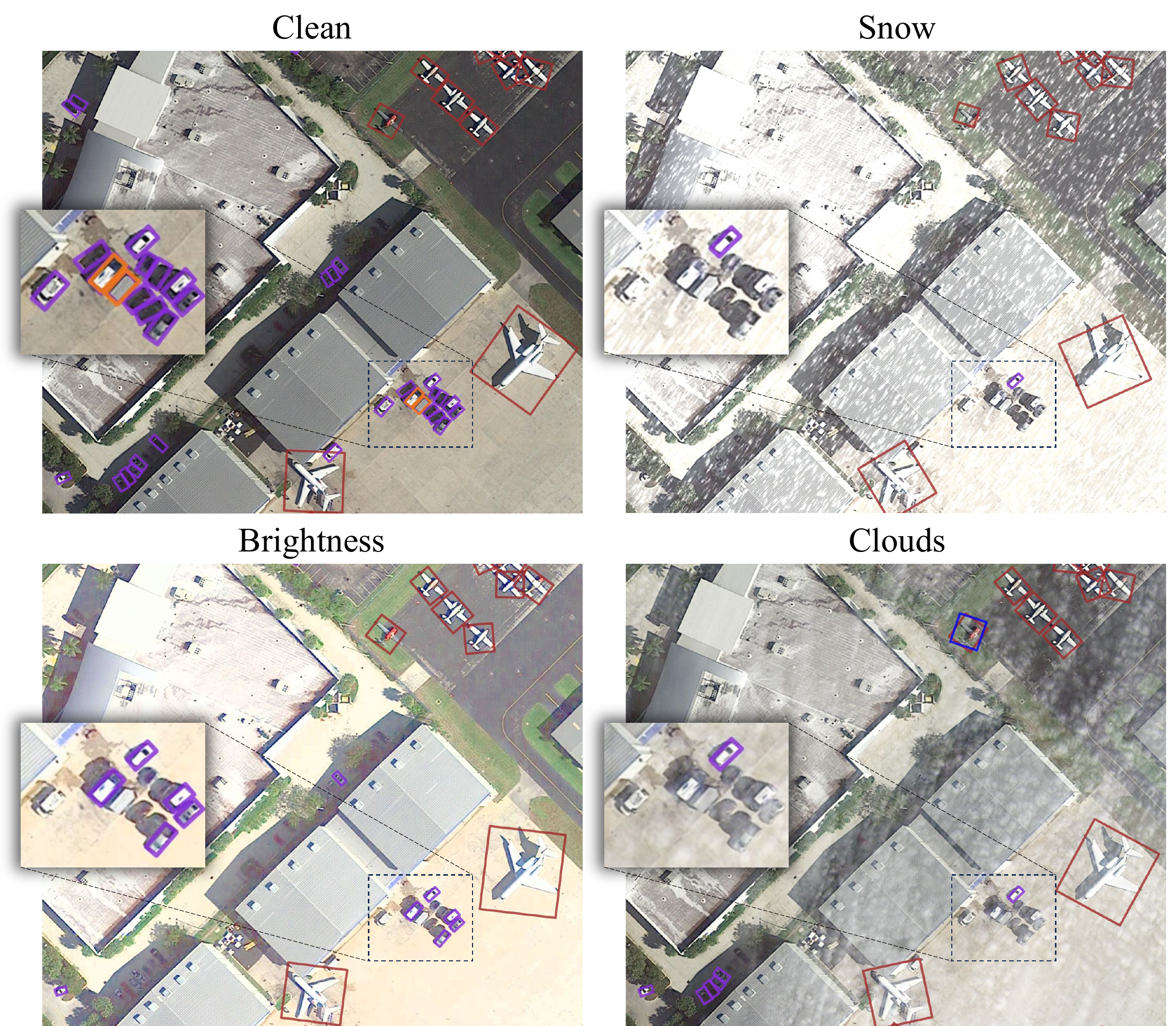}
    \caption{The object detection outcomes of RoI Transformer \cite{roitrans} on a clean test image and the same image subject to various corruptions, including Snow and Brightness with a severity of 3, as well as Clouds. The model was trained just on clean data without exposure to these corruptions and bounding boxes with colors representing different object classes are displayed only if their confidence scores are not less than 0.3. Original images are cropped and enlarged to make visualization better. Due to corruptions of the image, there is a notable increase in false detection and missed detection.}
    \label{fig:example}
\end{figure}

\IEEEpubidadjcol

Contrary to natural images, object detection on aerial images presents unique challenges with the bird's eye view, the highly complex backgrounds, the large variations in object scale, and the arbitrary orientations \cite{challenge1,challenge2}. Learning-based object detection models are very likely to have difficulty handling corrupted aerial images. For instance, the advanced RoI Transformer~\cite{roitrans} gets a lot of errors for the test image with some slight corruptions as shown in Figure~\ref{fig:example}. 
Although some efforts have been made to reduce the negative impact of image degradation, they tend to improve image quality through a preprocessing step~\cite{han2021methods}. There is still a lack of open-source benchmarks designed to assess the performance of models for object detection on corrupted data. The present datasets~\cite{dior,dota,sun2022fair1m} consisting of clean aerial images could not effectively measure the robustness of models.

Therefore, to fully evaluate the robustness of learning-based object detection models on aerial images, we build two benchmarks by applying sufficient image corruptions to a widely-used aerial object detection dataset named DOTA-v1.0~\cite{dota} in this paper. The first one consists of 19 common visual corruptions with five levels of severity from ImageNet-C~\cite{imagenetc}, while the second one is focused on a singular corruption category, designated as ``Clouds''.

Firstly, the visual corruptions~\cite{imagenetc} defined for natural images are widespread in aerial images of realistic scenes as well. For example, Gaussian noise in images arises during acquisition due to the level of illumination and the sensor's own temperature. Shot noise emerges as electronic noise stemming from the discrete nature of light. Defocus blur transpires when an image lacks precise focus, while Motion blur becomes apparent during swift camera movements. In general, we explore a total of 19 kinds of different corruptions that come from noise, blur, weather, and digital categories in one of the benchmarks termed DOTA-C.

Furthermore, it is worth noting that clouds are very common in aerial images but rarely appear in natural images. As a primary factor, cloud cover seriously affects the quality of aerial images and limits their applications~\cite{XU2019thin,liu2023thick}. Hence, the presence of clouds is a pivotal consideration in assessing the robustness of aerial object detection models. In accordance with this premise, we introduce a new corruption type called ``Clouds" for the second benchmark termed DOTA-Cloudy. Instead of synthetically generated clouds, the Clouds corruption is constructed by performing cloud transference from authentic cloudy satellite images to cloud-free clean images to reduce distortion.

Within the scope of these two benchmarks, we undertake a comprehensive evaluation of the robustness exhibited by various advanced object detection models. The detailed analysis and interpretation of experimental data are carried out. 
Moreover, we engage in an array of ablation experiments, encompassing factors such as alterations in backbone architectures, variations in the capacity within the same backbone, and the application of diverse data augmentation strategies. 
Our contributions can be summarized as follows: 

\begin{itemize}
   \item  We build two robustness benchmarks for object detection on aerial images called DOTA-C and DOTA-Cloudy respectively. The first consists of 19 common corruptions, and the second focuses on cloud cover as an essential aspect of assessing the quality of aerial images. %

   \item We evaluate the robustness of numerous representative object detection models on the proposed benchmarks, which could serve as a baseline for further research.
   
   \item We conduct a thorough analysis of the evaluation results and reveal that tested models not only have decreased performance but also display some new behaviors. We find that rotation-invariant modeling and enhanced backbone architectures can improve the robustness of models. Moreover, increasing the capacity of Transformer-based backbones can strengthen their robustness.  

\end{itemize}

\section{Related Work}

\subsection{Aerial Object Detection}
To address the unique characteristics of objects on aerial images such as arbitrary orientations and varying densities, the task of aerial object detection is often defined as oriented object detection, requiring identifying and locating objects with rotated bounding boxes.

Deep learning models like convolutional neural network (CNN) have been introduced to object detection methods since the proposal of R-CNN \cite{rcnn} and become mainstream recently.

For instance, based on features extracted by CNN, the rotation anchors~\cite{rotate} with extra angler parameters are designed to generate rotation region proposals for further classification and regression in a two-stage detection framework~\cite{fasterrcnn}. The increased number of anchors results in a lot of extra computational effort. RoI Transformer \cite{roitrans} was then proposed to deal with the rotation prediction by learning the transformation between the horizontal region of interest (RoI) and rotated RoI through fully connected layers. Although detection results are improved, feature maps taken by the rotated RoI are not real rotation-invariant, which restricts the accuracy of oriented object detection. 
ReDet~\cite{redet} later employed rotation-equivariant CNNs and rotation-invariant RoI Align module to overcome this limitation and extract genuine rotation-invariant features. 
Besides two-stage detectors, another framework of object detection does not require detached proposal generation. The one-stage detectors directly detect objects in one pass, achieving faster and more efficient processing. 
Over the years, many one-stage detectors specifically designed for aerial object detection have emerged. For instance, $\mathrm{S^2A}$-Net \cite{s2anet} introduced the Anchor Refinement Network and Alignment Convolution Layer to refine initial anchors and align features according to the rotated anchor frames. $\mathrm{R^3Det}$ \cite{r3det} adopts a progressive regression approach to accommodate objects of varying densities and aspect ratios. Some subsequent works for aerial object detection also made remarkable contributions in improving the accuracy of models and reducing the detectors' number of parameters \cite{orientedrcnn,sasm,reppoints,oan,sfrnet,psc}. For example, SFRNet \cite{sfrnet} advances fine-grained oriented object recognition through separate feature refinement. OAN \cite{oan} can enhance both efficiency and accuracy by focusing on relevant patches and integrate seamlessly with existing detectors.

\subsection{Robustness in Object Detection}
Most of the researches evaluate the robustness of object detection models by introducing corruptions into clean datasets. These corruptions can be broadly divided into adversarial perturbations~\cite{adversarial1,adversarial2,adversarial3} and common corruptions~\cite{corruptions1,corruptions2,corruptions3,imagenetc}. In this paper, we concentrate on assessing the impact of corruptions on the models' performance, given their higher occurrence in real-world data in comparison to adversarial perturbations.

There are currently some robustness benchmarks in the field of natural image object detection. For example, Hendrycks and Dietterich~\cite{imagenetc} introduced 19 common corruptions and applied them to the ImageNet dataset \cite{imagenet}, obtaining ImageNet-C.
Michaelis et al. \cite{autonomousdriving} utilized the methodologies outlined in ImageNet-C to establish benchmark datasets including Pascal-C, COCO-C, and Cityscapes-C. Additionally, there are also quite a few studies on the robustness of natural object detection models. In \cite{corruptions1}, Dodge \textit{et al.} proved that CNNs are susceptible to quality distortions, particularly to blur and noise. Geirhos \textit{et al.} \cite{corruptions2} confirmed human vision system is more robust than CNNs on object recognition under several image degradations. The research of Azulay \textit{et al.} \cite{corruptions3} showed that CNNs' accuracy would drop dramatically because of slight geometric transformations. Unfortunately, the domain of aerial object detection suffers from a scarcity of both robustness benchmarks and research on robustness. Therefore, we construct two new robustness benchmarks and investigate the robustness of aerial object detection models.

\subsection{Corruption Robustness Improvement}
There are several methods to improve models' accuracy on corrupted images. One of them is to remove the corruptions from images. He \textit{et al.} \cite{tradition3} used ``dark channel prior'' to remove haze from images and got good results. In \cite{restoration}, SwinIR was proposed to restore high-quality images from downscaled, noisy, and compressed images, and it demonstrated better performance than state-of-the-art methods in the case of reduced parameter quantity. However, both traditional methods \cite{tradition1,tradition2,tradition3} and deep learning-based methods \cite{denoise1,denoise2,denoise3,restoration} can only handle one or a few image degradations. These approaches can't be generalized to other types of distortions. Another method is to add corrupted data into the models' training set. Vasiljevic \textit{et al.} \cite{blur} observed that CNNs can enhance their performance in object detection for data affected by the same corruption, by undergoing fine-tuning on blurred images. Geirhos \textit{et al.} \cite{corruptions2} revealed that models, when trained on specific corruption types, could outperform the human vision system on those exact corruption types. However, these models exhibited notably limited generalization capabilities when evaluated on different corruption types. Hendrycks and Dietterich~\cite{imagenetc} reported several methods to improve models' robustness: Histogram Equalization, Multiscale Networks, Larger Networks, and so on. In this work, we conduct a series of ablation experiments to figure out how to improve the robustness of aerial object detection models.

\section{Robustness Benchmarks for Aerial Object Detection}

We measure the robustness of object detection models on aerial images by establishing two benchmark datasets in this section. Each dataset is composed of clean images for model training and corrupted images for testing.

The original clean aerial data comes from the DOTA dataset~\cite{dota}, which is one of the most widely used aerial object detection datasets. 
It exhibits a great abundance of object instances, random yet uniform directions, diverse classification categories, and intricate aerial scenes. Moreover, the scenes within the DOTA dataset~\cite{dota} align closely with real-world scenarios, making it more valuable for the development of applications in practical settings. To be specific, we utilize DOTA-v1.0~\cite{dota}, which labels 15 classes of objects including ship, storage tank, baseball diamond, tennis court, basketball court, ground track field, harbor, bridge, large vehicle, small vehicle, helicopter, roundabout, soccer ball field, and basketball court. The DOTA-v1.0 dataset~\cite{dota} has 1411 training images, 458 validation images, and 937 test images.
Corruptions are only imposed on the test set. It should be noted that the approach in which corruptions are implemented can also be extended to other datasets.

In the following, we present details about the two constructed datasets for robust aerial object detection along with their evaluation metrics.

\subsection{Common Corruptions} 
\begin{figure*}[htb!]
    \centering
    \includegraphics[width=\textwidth]{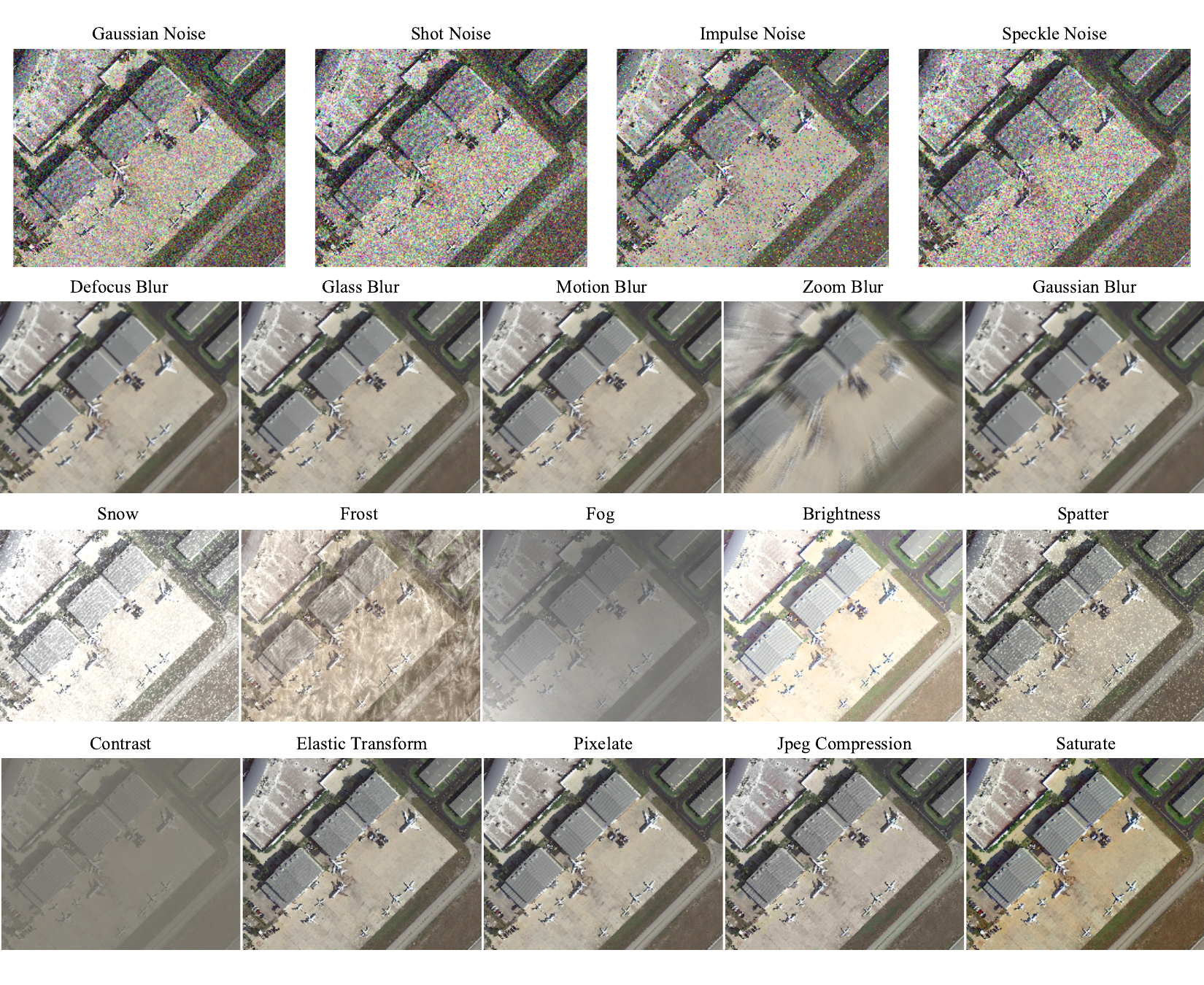}
    \caption{Visualization of all types of corruptions with severity 3 in the DOTA-C dataset on a sampled aerial image. Each row shows one of the four categories.}
    \label{fig:dotac}
\end{figure*}

\begin{figure}[h]
    \centering
    \includegraphics[width=\linewidth]{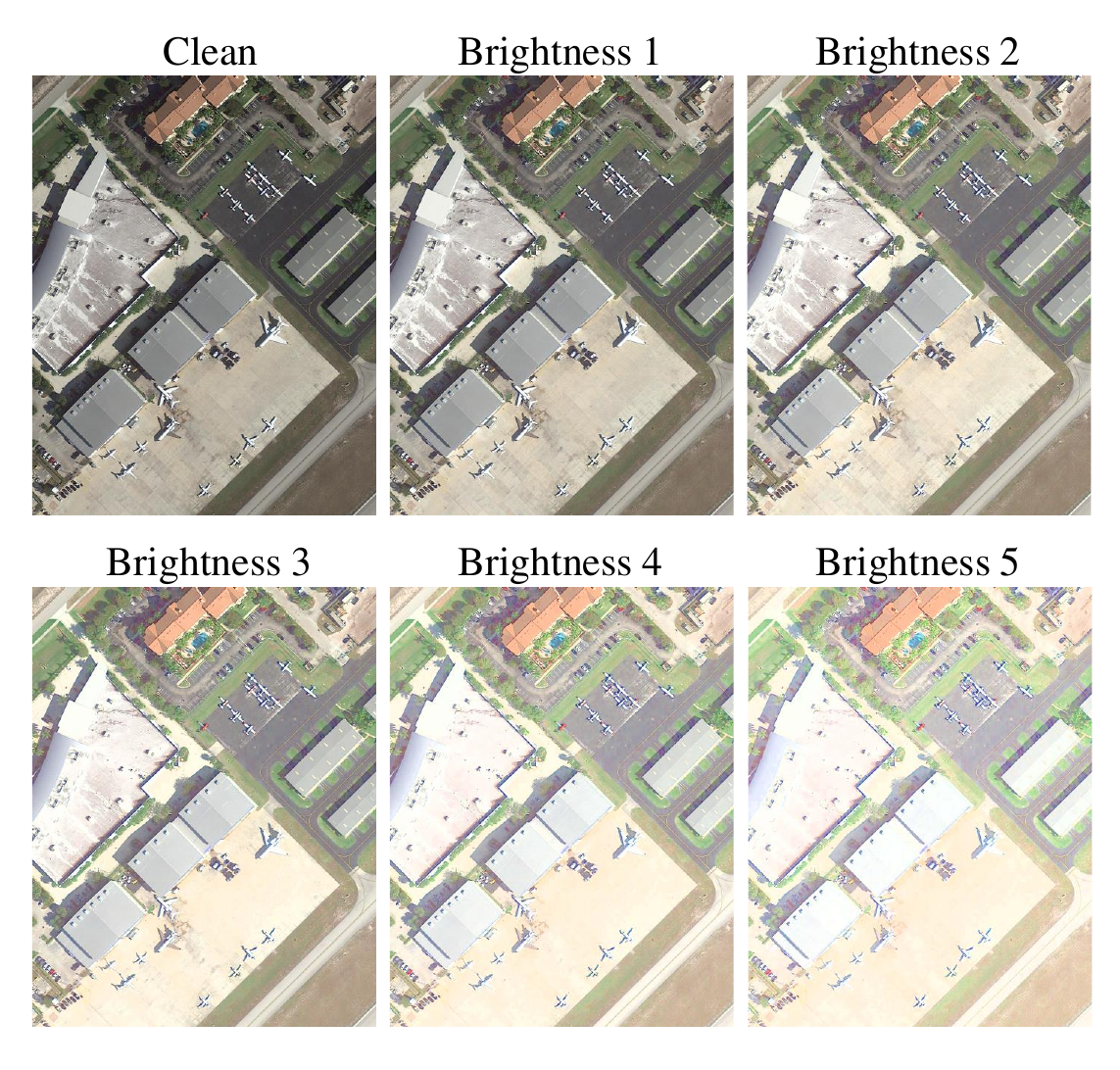}
    \caption{A clean image and the same image corrupted with ``Brightness" of levels 1, 2, 3, 4, and 5. From 1 to 5, the imposed corruptions increased progressively.}
    \label{fig:brightness}
\end{figure}

The DOTA-C dataset measures the common corruptions derived from the ImageNet-C dataset \cite{imagenetc}.
We use all the corruptions in the ImageNet-C dataset, which contains 19 types of corruptions and each one has 5 levels to assess its severity. All corruptions, divided into four categories are listed as: Noise: Gaussian, Shot, Impulse, and Speckle; Blur: Defocus, Glass, Motion, Zoom, and Gaussian; Weather: Snow, Frost, Fog, Brightness, and Spatter; Digital: Contrast, Elastic transform, Pixelate, JPEG compression, and Saturate. In Figure \ref{fig:dotac}, we illustrate all 19 types of corruptions. In Figure \ref{fig:brightness}, we show the five different severity levels for one of the corruptions called ``Brightness". On the whole, the collection of corruption types for all levels are applied to test images of DOTA-v1.0 dataset~\cite{dota} to make up the DOTA-C dataset.

Following the setting of the DOTA dataset~\cite{dota}, we apply $\mathrm{AP_{50}}$ as the evaluation metric, which stands for the mean of Average Precision over all detected object classes at a threshold of 50\% for Intersection over Union (IoU).
For each one corruption type \(c\) at severity level \( s\), we calculate its quantitative detection results and express it as \({\rm AP}^{c, s}_{50}\).
Our evaluation of models is based on the mean performance under corruption ($\mathrm{mPC}$) \cite{autonomousdriving}, which is formulated as follows:

\begin{equation}
{\rm mPC}=\frac{1}{~ {\rm N_{c}}} \sum_{c=1}^{\rm N_{c}} \frac{1}{~{\rm N_{s}}} \sum_{s=1}^{\rm N_{s}}{\rm AP}^{c, s}_{50}\,.
\label{equation:4}
\end{equation}

The values $\mathrm{N_c}$ = 19 and $\mathrm{N_s}$ = 5 denote the number of corruption types and severity levels, respectively. It is important to note that a higher $\mathrm{mPC}$ does not inherently signify superior robustness of the model, since its performance might undergo a swifter deterioration in the presence of corruptions compared to a model with a lower $\mathrm{mPC}$. Consequently, to quantify the performance degradation induced by corruptions, relative performance under corruption ($\mathrm{rPC}$) \cite{autonomousdriving} is put forward and defined as below:

\begin{equation}
{\rm rPC}=\frac{\rm mPC}{\rm AP^{\text {clean}}_{50}}\,,
\label{equation:5}
\end{equation}
where $\mathrm{AP}^{\text {clean}}_{50}$ represents a model's $\mathrm{AP}_{50}$ on the clean test set. A higher {\rm rPC} means the model's performance is less affected by corruptions, showing better robustness.

\subsection{Cloud Cover}

\begin{figure}[h!]
    \centering
    \includegraphics[width=\linewidth]{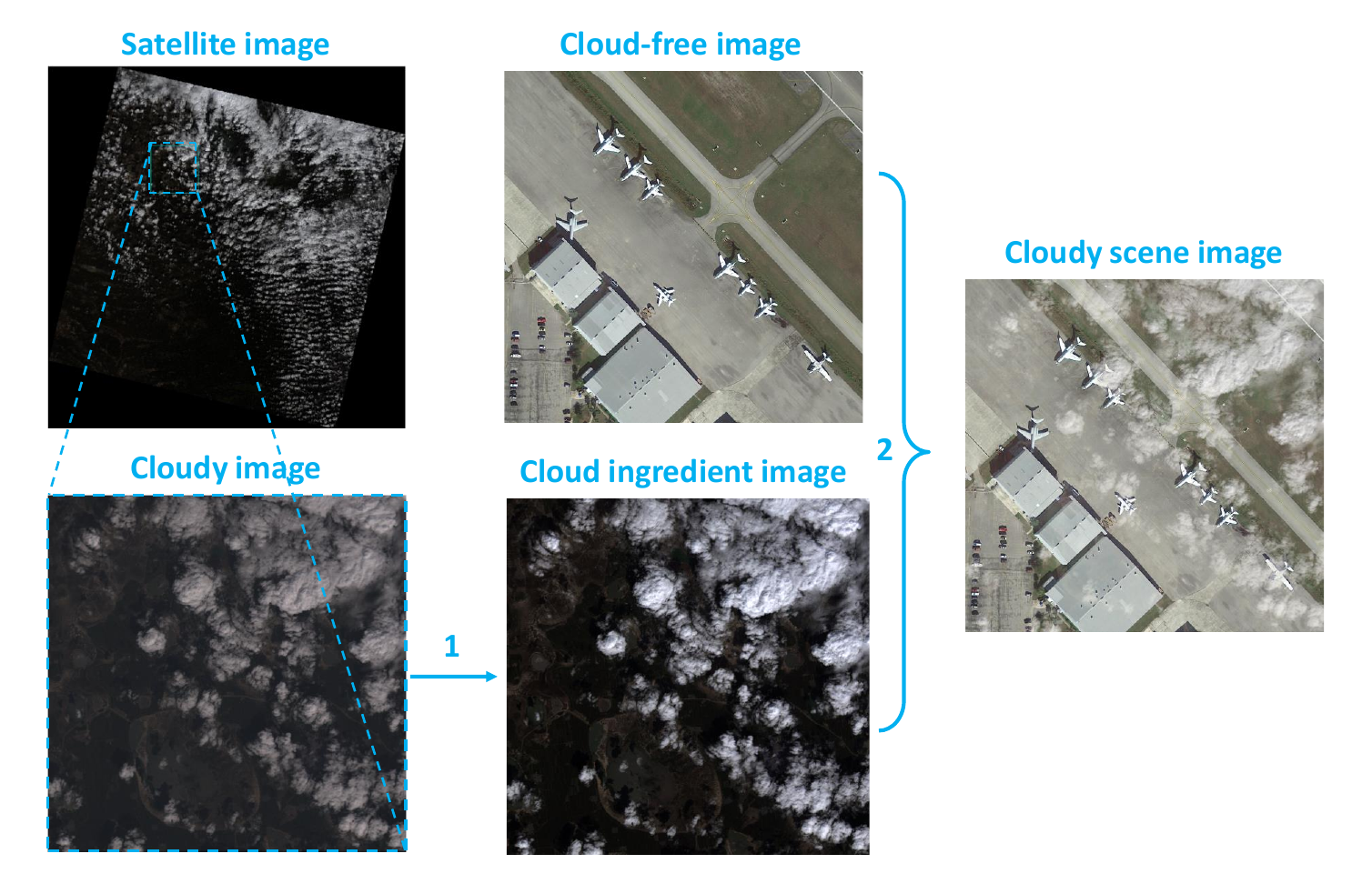}
    \caption{The cloud-corrupted image generation scheme for the DOTA-Cloudy dataset. After getting a cropped cloudy image from the original satellite image, there are two steps to apply clouds to a cloud-free image. The number ``1'' represents the first step ``Cloud Self-Subtraction", followed by the second step``Cloud Addition-to-Scene" marked in the number ``2''.}
    \label{fig:clouds}
\end{figure}

The DOTA-Cloudy dataset emphasizes the robustness of object detection models against the corruption caused by cloud cover. It is hard to directly get real high-resolution aerial images covered by clouds along with their corresponding object labels. Therefore, the Clouds corruption is built synthetically. However, the existing cloud images generated based on generative adversarial networks \cite{gan} and their derivative models usually have difficulty in simulating clouds with natural shapes and appearances. To enhance the visual consistency with the real scene, we choose to use the ``Cloudy Image Arithmetic" approach \cite{clouds}, which can transfer clouds from cloudy images to clean target images. As shown in Figure~\ref{fig:clouds}, the whole process of generating cloud-corrupted images mainly consists of two steps: the ``Cloud Self-Subtraction" and the ``Cloud Addition-to-Scene". 

The first step is to extract clouds from a cloudy image: \par

\begin{equation}
I_{dc}(m, n)={\rm ReLU}\left[(I_{cs}-\Gamma \mathbf{1})(m, n)\right]\,.
\label{equation:1}
\end{equation}

In this context, $I_{cs}$ denotes a cloudy image of size $M \times N$ that covers both cloud-free and cloudy sub-areas. $\mathbf{1}$ presents an $M \times N$ matrix with all one elements. A pixel \((m,n)\) with intensity less than a threshold value $\Gamma$ is categorized into the background. The purpose of the ReLU function is to make all values greater than 0 unchanged and set all values less than 0 to 0. $I_{dc}$ is a degraded cloud representation.

The process in \eqref{equation:1} degrades both the background pixels and cloud pixels. Therefore, to get the cloud ingredient image ($I_{ci}$), we need to give it a compensation coefficient:

\begin{equation}
I_{ci}=\frac{\sum_{m=1}^{M} \sum_{n=1}^{N} I_{cs}(m, n) \mathbbm{1}\left[I_{dc}(m, n) \neq 0\right]}{\sum_{m=1}^{M} \sum_{n=1}^{N} I_{dc}(m, n)} I_{dc}\,,
\label{equation:2}
\end{equation}
where $\mathbbm{1}$ denotes an indicator function, assigned the value of 1 when its variable holds true and 0 otherwise. 

\begin{figure}[h!]
    \centering
    \includegraphics[width=\linewidth]{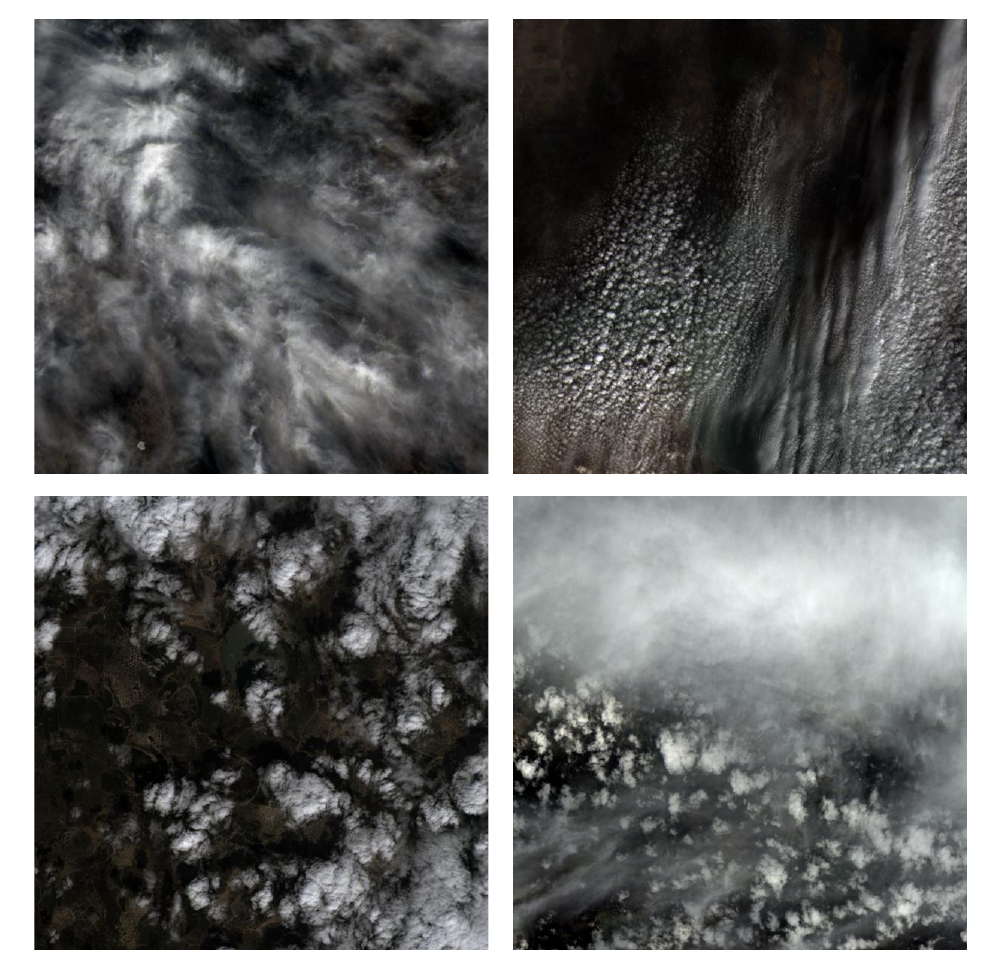}
    \caption{Examples in the set of collected cloudy scene images.}
    \label{fig:clouds_example}
\end{figure}

In order to get cloudy images, we acquire 10 satellite images from USGS Landsat 8 Collection 2 Tier 1 Raw Scenes (Landsat-8 image courtesy of the U.S. Geological Survey) and select the three bands ``B4'', ``B3'' and ``B2'' to make them true color images. Subsequently, we crop these images and finally get 30 cloudy scene images with the size of $1024\times1024$ pixels. Figure \ref{fig:clouds_example} shows some examples with clouds of varying thicknesses and structures. This new set of cloudy images contains various kinds of clouds, which can well simulate the problem of cloud cover in actuality. 

The second step is to synthesize the cloudy scene image by adding clouds from the cloud ingredient image \(I_{ci}\) to a clean image. It can be formulated as follows:

\begin{equation}
I_{scs}=I_{cfl}(G\mathbf{1}-I_{ci})+A I_{ci}\,,
\label{equation:3}
\end{equation}
where $I_{scs}$ refers to the synthesized cloudy scene image and $I_{cfl}$ refers to the cloud-free land image. \( G\) presents the maximum gray-level value, which is 255 here. \( A\) means atmospheric light and is set to 0.95 as referred to~\cite{light}. \par

According to \eqref{equation:2} and \eqref{equation:3}, we construct the DOTA-Cloudy dataset by transferring clouds from the collected set of cloudy scene images to clean images from the test set of the DOTA-v1.0 dataset~\cite{dota} in batches.

Due to the processing approach, ``Clouds'' in the DOTA-Cloudy dataset can not be easily graded. So the evaluation metrics are denoted as $\mathrm{AP}^{\text {clouds}}_{50}$ and $\mathrm{rPC}_{\text {clouds}}$. They are defined as follows:

\begin{equation}
{\rm rPC}_{\text {clouds}}=\frac{\rm AP^{\text {clouds}}_{50}}{\rm AP^{\text {clean}}_{50}}\,,
\label{equation:6}
\end{equation}
where $\mathrm{AP}^{\text {clouds}}_{50}$ means a model's $\mathrm{AP}_{50}$ on data corrupted with clouds and $\mathrm{rPC}_{\text {clouds}}$ presents the relative performance degradation of a model caused by clouds. 

\section{Experiments}

\subsection{Model Evaluation}

In general, we assess eleven object detection methods on the benchmarks introduced in this study.
Among them, two-stage models include Rotated Faster R-CNN \cite{fasterrcnn}, RoI Transformer \cite{roitrans}, Oriented R-CNN \cite{orientedrcnn}, ReDet \cite{redet}, SFRNet \cite{sfrnet}, and OAN \cite{oan}. They first generate region proposals and then predict object classification and localization using CNNs. One-stage models include Rotated RetinaNet \cite{retinanet}, Rotated FCOS \cite{fcos}, $\mathrm{R^3Det}$ \cite{r3det}, $\mathrm{S^2A}$-Net \cite{s2anet}, and PSC \cite{psc}. They involve directly extracting features using CNNs to predict object classification and localization. All these models' backbone is ResNet50 \cite{resnet} with Feature Pyramid Networks (FPN) \cite{fpn}, except for ReDet, which employs ReResNet50 with ReFPN \cite{redet}. OAN~\cite{oan} and PSC~\cite{psc} are both implemented as modules integrated to RoI Transformer~\cite{roitrans} and RetinaNet~\cite{retinanet} respectively.

\subsubsection{Implementation details}
The entirety of the training and testing procedures are executed through the utilization of the MMRotate toolbox \cite{mmrotate}, with the exception of SFRNet \cite{sfrnet}, OAN \cite{oan}, and PSC \cite{psc}. Since they are not integrated into MMRotate yet, we train and test them using their respective official codes. Following the previous works~\cite{roitrans,redet}, we use the training set and the validation set for training and the test set for testing. 
Every model has been trained on a machine with 4 GPUs with batch size 8 (2 per GPU), standard hyperparameters, and the same image normalization configurations. 

During the training phase, we split the original images to $1024\times1024$ pixels and set the value of overlap to 200 pixels. 
Before the training procedure, a series of preprocessing steps are administered to the images. These include resizing, random horizontal flipping, and normalization, after which the images are transformed into RGB format. The training phase spans 12 epochs for all models, adhering to a consistent learning rate policy. This policy encompasses a linear warmup stage, with the learning rate diminishing to one-tenth of its original value after the eighth and eleventh epochs.

During the testing phase, the test images undergo a two-step process: initial corruption and subsequent split. After the alteration induced by corruptions, images are cropped using the same strategy as the training phase. Subsequently, resizing and normalization are carried out, culminating in the conversion of the images to RGB format.

\subsubsection{Results and analysis}
For the common corruptions in the DOTA-C dataset, models' $\mathrm{AP}^{\text {clean}}_{50}$, $\mathrm{mPC}$ and $\mathrm{AP}_{50}$ for each corruption type averaged over all severity levels are displayed in Table \ref{table:mpc}. Due to the limited space, we have adopted an abbreviated form to denote the name of corruption type, such as ``Ga." indicates the Gaussian Noise, ``Im.'' indicates the Impulse Noise, and so forth.

\begin{table*}[]
\begin{center}
\caption{Quantitative evaluation results of tested models on the DOTA-C dataset. Starting from the fourth column, the values in each column represent $\mathrm{AP}_{50}$ of the corresponding models for the single corruption across five severity levels.}
\label{table:mpc}
\resizebox{\linewidth}{!}{
\begin{tabular}{lccccccccccccccccccccc}
\hline
          & \multicolumn{1}{l}{}   & \multicolumn{1}{l}{}    & \multicolumn{1}{l}{} & \multicolumn{2}{c}{Noise} & \multicolumn{1}{l}{}     & \multicolumn{1}{l}{} & \multicolumn{1}{l}{} & Blur & \multicolumn{1}{l}{} & \multicolumn{1}{l}{}      & \multicolumn{1}{l}{} & \multicolumn{1}{l}{} & Weather & \multicolumn{1}{l}{} & \multicolumn{1}{l}{}      & \multicolumn{1}{l}{} & \multicolumn{1}{l}{} & Digital & \multicolumn{1}{l}{} & \multicolumn{1}{l}{} \\ \hline
Model     & \multicolumn{1}{|c}{$\mathrm{AP}^{\text {clean}}_{50}$}  & \multicolumn{1}{c|}{mPC} & Ga. & Shot & Im.   & \multicolumn{1}{c|}{Spec.} & De.  & Glass  & Mo.  & Zoom   & \multicolumn{1}{c|}{Ga.} & Snow  & Frost  & Fog  & Br.   & \multicolumn{1}{c|}{Spat.} & Co.  & El.  & Pixel  & JPEG  & Sa.         \\ \hline
Rot. Faster \cite{fasterrcnn} & \multicolumn{1}{|c}{73.4} & \multicolumn{1}{c|}{38.9}   & 20.2  & 19.7 & 17.7  & \multicolumn{1}{c|}{27.6}   & 40.5  & 46.4  & 40.6   & 14.1   & \multicolumn{1}{c|}{43.0}   & 24.3  & 46.2  & 49.3   & 63.1   & \multicolumn{1}{c|}{46.7}   & 42.4  & 33.2  & 53.1   & 50.4   & 60.7 \\ 
RoI Trans. \cite{roitrans}  & \multicolumn{1}{|c}{76.1} & \multicolumn{1}{c|}{39.9}   & 19.8  & 20.2 & 17.8  & \multicolumn{1}{c|}{29.1}   & 41.1  & 48.8  & 42.6   & 14.7   & \multicolumn{1}{c|}{44.0}   & 26.5  & 47.1  & 49.2   & 63.5   & \multicolumn{1}{c|}{49.4}   & 42.5  & 35.0  & 53.6   & 51.5   & 62.3 \\ 
Ori. R-CNN \cite{orientedrcnn}  & \multicolumn{1}{|c}{75.7} & \multicolumn{1}{c|}{40.7}   & 21.7  & 21.7 & 18.7  & \multicolumn{1}{c|}{30.3}   & 41.9  & 49.0  & 42.3   & 14.8   & \multicolumn{1}{c|}{44.3}   & 25.6  & 48.7  & 51.5   & 65.6   & \multicolumn{1}{c|}{48.5}   & 43.2  & 34.9  & 55.5   & 50.7   & 63.4 \\ 
ReDet \cite{redet}  & \multicolumn{1}{|c}{76.7} & \multicolumn{1}{c|}{45.9}   & 24.7  & 24.6 & 22.4  & \multicolumn{1}{c|}{34.3}   & 50.3  & 53.6  & 48.3   & 18.1   & \multicolumn{1}{c|}{53.2}   & 35.3  & 58.3  & 63.1   & 70.5   & \multicolumn{1}{c|}{52.0}   & 54.3  & 33.2  & 59.4   & 52.4   & 64.9 \\ 
SFRNet \cite{sfrnet}  & \multicolumn{1}{|c}{75.9} & \multicolumn{1}{c|}{41.3}   & 22.0  & 22.2 & 19.6  & \multicolumn{1}{c|}{30.4}   & 42.6  & 49.4  & 43.6   & 14.8   & \multicolumn{1}{c|}{45.4}   & 28.0  & 48.7  & 51.5   & 66.1   & \multicolumn{1}{c|}{49.3}   & 44.0  & 35.2  & 55.6   & 52.5   & 63.5 \\
OAN \cite{oan}  & \multicolumn{1}{|c}{73.9} & \multicolumn{1}{c|}{40.0}   & 19.4  & 20.1 & 17.2  & \multicolumn{1}{c|}{28.6}   & 41.6  & 49.0  & 43.8   & 14.6   & \multicolumn{1}{c|}{44.4}   & 26.0  & 47.6  & 50.1   & 64.1   & \multicolumn{1}{c|}{48.8}   & 42.7  & 34.5  & 53.9   & 51.5   & 61.8 \\ \hline
Rot. Retina. \cite{retinanet}  & \multicolumn{1}{|c}{68.4} & \multicolumn{1}{c|}{37.3}   & 20.0  & 19.7 & 16.9  & \multicolumn{1}{c|}{26.7}   & 40.5  & 45.8  & 39.6   & 14.0   & \multicolumn{1}{c|}{43.3}   & 23.3  & 45.2  & 47.9   & 59.4   & \multicolumn{1}{c|}{42.9}   & 40.3  & 31.5  & 48.0   & 46.4   & 58.1 \\ 
Rot. FCOS \cite{fcos}  & \multicolumn{1}{|c}{71.3} & \multicolumn{1}{c|}{38.9}   & 20.6  & 20.5 & 18.7  & \multicolumn{1}{c|}{27.6}   & 41.3  & 46.8  & 39.6   & 14.5   & \multicolumn{1}{c|}{43.7}   & 26.1  & 46.6  & 50.7   & 61.2   & \multicolumn{1}{c|}{45.8}   & 43.8  & 31.6  & 51.4   & 48.3   & 59.6 \\
$\mathrm{R^3Det}$ \cite{r3det}  & \multicolumn{1}{|c}{69.8} & \multicolumn{1}{c|}{37.8}   & 19.9  & 19.6 & 17.3  & \multicolumn{1}{c|}{27.4}   & 38.6  & 44.3  & 38.0   & 14.4   & \multicolumn{1}{c|}{42.0}   & 24.8  & 46.5  & 48.6   & 61.1   & \multicolumn{1}{c|}{43.8}   & 41.8  & 31.8  & 51.7   & 47.1   & 59.4 \\
$\mathrm{S^2A}$-Net \cite{s2anet}  & \multicolumn{1}{|c}{73.9} & \multicolumn{1}{c|}{39.8}   & 18.6  & 18.6 & 15.7  & \multicolumn{1}{c|}{26.3}   & 42.3  & 48.4  & 41.2   & 15.1   & \multicolumn{1}{c|}{44.9}   & 28.7  & 49.7  & 53.2   & 64.0   & \multicolumn{1}{c|}{46.5}   & 45.0  & 33.8  & 50.9   & 49.9   & 62.7 \\ 
PSC \cite{psc}  & \multicolumn{1}{|c}{71.9} & \multicolumn{1}{c|}{37.9}   & 18.3  & 18.2 & 16.0  & \multicolumn{1}{c|}{25.3}   & 41.5  & 46.0  & 40.6   & 14.4   & \multicolumn{1}{c|}{44.7}   & 23.8  & 46.0  & 49.9   & 61.3   & \multicolumn{1}{c|}{44.4}   & 42.3  & 32.8  & 48.0   & 46.8   & 59.2 \\ \hline
\end{tabular}
}
\end{center}
\end{table*}

\begin{figure*}[]
    \centering
    \includegraphics[width=\textwidth]{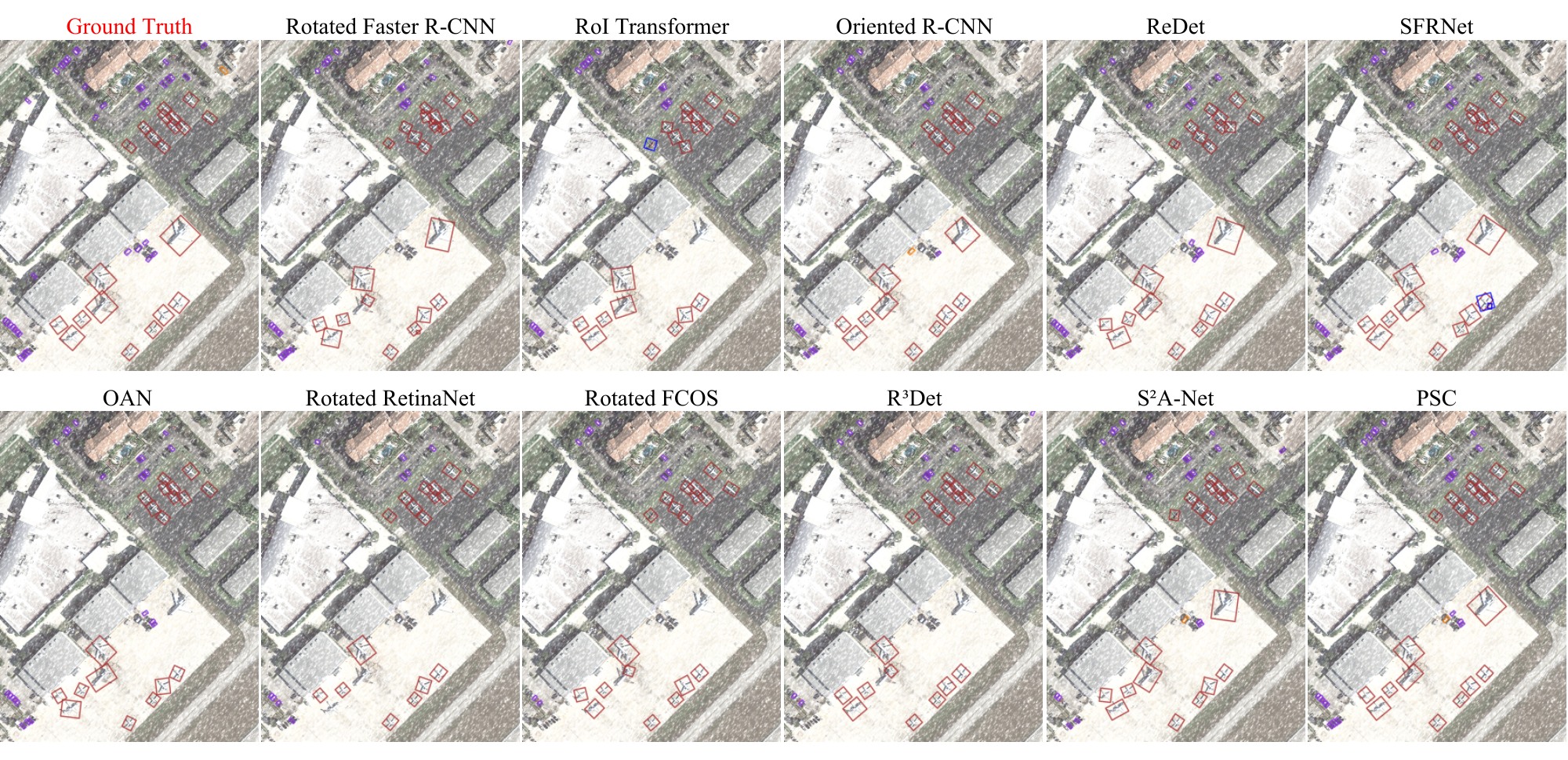}
    \caption{Object detection results of tested models on an aerial image corrupted with Snow at severity level 3. These images only display bounding boxes with a confidence score of no less than 0.3. In the image displaying the ground truth, red, purple, and orange represent the object categories of plane, small vehicle, and large vehicle, respectively.}
    \label{fig:experiment}
\end{figure*}

First of all, the large numerical gap between the values of $\mathrm{mPC}$ and $\mathrm{AP}^{\text {clean}}_{50}$ reveals that all tested models experience a significant performance degradation on the corrupted data. Models trained on regular clean data have restricted stability against unknown image corruptions. To make the performance of the models more intuitive, we show the object detection results on data corrupted with Snow at severity level 3 in Figure \ref{fig:experiment}. We can clearly observe that, compared to the ground truth, all models exhibit instances of missed detections and false positives, likely due to the insufficient ability to handle small objects. In addition, there are some models with deviations in the localization of bounding boxes, and such deformations may be caused by inaccuracies in feature extraction or spatial transformation by the models. Meanwhile, by observing the $\mathrm{AP}_{50}$ of the models under various corruptions, it can be seen that the performance of tested models varies significantly for different types. Among all types, ``Brightness" has a minimal impact on models' performance, whereas ``Zoom blur" degrades it the most. Their $\mathrm{AP_{50}}$ scores differ by more than 40\%. 

Secondly, comparing results among the tested models in Table~\ref{table:mpc}, the two-stage detectors shown on the first few rows of Table~\ref{table:mpc} generally get better performance than one-stage detectors on both clean and corrupted test set. Specifically, $\mathrm{S^2A}$-Net \cite{s2anet} outperforms other one-stage models, and ReDet \cite{redet} exhibits superior performance to other two-stage models with the highest $\mathrm{AP}_{50}^{{\rm clean}}$ and \({\rm mPC}\) scores. However, it can be noticed that the models performed superior on clean data do not guarantee better detection results on corrupted data. For example, the $\mathrm{AP}_{50}^{{\rm clean}}$ of RoI Transformer \cite{roitrans} is 76.1\% and Oriented R-CNN \cite{orientedrcnn} is 75.7\%. But their $\mathrm{mPC}$ values rank in a reversed order as 39.9\% and 40.7\%. A similar phenomenon can be observed in Rotated FCOS~\cite{rotate} and PSC~\cite{psc} too.

We later conduct a quantitative evaluation and analysis of the degradation in models' performance on image corruptions. The evaluation results measured in $\mathrm{rPC}$ can be seen in Table \ref{table:corruption_results}. Among the two-stage models, the performance of RoI Transformer \cite{roitrans} degrades the most and gets the lowest $\mathrm{rPC}$. Rotated RetinaNet \cite{retinanet} gets the best results in one-stage models, while ReDet \cite{redet} achieves the highest $\mathrm{rPC}$ among all models. Except for ReDet~\cite{redet}, the vast majority of tested models have similar scores of $\mathrm{rPC}$. It is indicated that in terms of performance stability, the difference in detection frameworks (one-stage or two-stage) has no obvious influence. 

\begin{table}[]
\begin{center}
\caption{Relative performance evaluation of tested models on the DOTA-C dataset.}
\label{table:corruption_results}
\resizebox{\linewidth}{!}{
\begin{tabular}{l|lllll}
\hline
Model  & $\mathrm{rPC(\%)}$ & $\mathrm{rPC}_{\text {noise}}(\%)$   & $\mathrm{rPC}_{\text {blur}}(\%)$  & $\mathrm{rPC}_{\text {weather}}(\%)$  & $\mathrm{rPC}_{\text {digital}}(\%)$ \\ \hline
Rot. Faster \cite{fasterrcnn} & \multicolumn{1}{c}{53.01} & \multicolumn{1}{c}{29.01} & \multicolumn{1}{c}{50.31} & \multicolumn{1}{c}{62.56} & \multicolumn{1}{c}{65.38}  \\ 
RoI Trans. \cite{roitrans} & \multicolumn{1}{c}{52.46} & \multicolumn{1}{c}{28.55} & \multicolumn{1}{c}{50.24} & \multicolumn{1}{c}{61.95} & \multicolumn{1}{c}{64.35}  \\ 
Ori. R-CNN \cite{orientedrcnn} & \multicolumn{1}{c}{53.71} & \multicolumn{1}{c}{30.52} & \multicolumn{1}{c}{50.84} & \multicolumn{1}{c}{63.39} & \multicolumn{1}{c}{65.45} \\ 
ReDet \cite{redet}  & \multicolumn{1}{c}{59.90} & \multicolumn{1}{c}{34.55} & \multicolumn{1}{c}{58.27} & \multicolumn{1}{c}{72.81} & \multicolumn{1}{c}{68.91} \\ 
SFRNet \cite{sfrnet}  & \multicolumn{1}{c}{54.39} & \multicolumn{1}{c}{31.04} & \multicolumn{1}{c}{51.59} & \multicolumn{1}{c}{64.18} & \multicolumn{1}{c}{66.10} \\ 
OAN \cite{oan}  & \multicolumn{1}{c}{54.08}  & \multicolumn{1}{c}{28.84} & \multicolumn{1}{c}{52.34}  & \multicolumn{1}{c}{64.00} & \multicolumn{1}{c}{66.10} \\ \hline
Rot. Retina. \cite{retinanet} & \multicolumn{1}{c}{54.57}  & \multicolumn{1}{c}{30.40} & \multicolumn{1}{c}{53.55} & \multicolumn{1}{c}{63.93} & \multicolumn{1}{c}{65.55} \\
Rot. FCOS \cite{fcos}  & \multicolumn{1}{c}{54.50} & \multicolumn{1}{c}{30.67} & \multicolumn{1}{c}{52.13} & \multicolumn{1}{c}{64.62} & \multicolumn{1}{c}{65.85} \\ 
$\mathrm{R^3Det}$ \cite{r3det} & \multicolumn{1}{c}{54.14} & \multicolumn{1}{c}{30.14} & \multicolumn{1}{c}{50.80} & \multicolumn{1}{c}{64.38} & \multicolumn{1}{c}{66.43} \\ 
$\mathrm{S^2A}$-Net \cite{s2anet} & \multicolumn{1}{c}{53.81} & \multicolumn{1}{c}{26.83} & \multicolumn{1}{c}{51.96} & \multicolumn{1}{c}{65.50} & \multicolumn{1}{c}{65.57} \\ 
PSC \cite{psc} & \multicolumn{1}{c}{52.67}  & \multicolumn{1}{c}{27.01} & \multicolumn{1}{c}{52.10} & \multicolumn{1}{c}{62.72} & \multicolumn{1}{c}{63.73} \\ \hline
\end{tabular}}
\end{center}
\end{table}

We also conduct a comprehensive investigation into the influence of 4 different corruption categories. As shown in Table \ref{table:corruption_results}, the $\mathrm{rPC_{\text{noise}}}$, $\mathrm{rPC}_{\text{blur}}$, $\mathrm{rPC}_{\text{weather}}$, and $\mathrm{rPC}_{\text{digital}}$ represent the average value of relative performance for a few types across five severity levels under each category denoted by subscripts. It is evident that weather corruptions and digital corruptions exhibit minimal impact on the models, while noise corruptions result in the poorest performance. Among the considered models, ReDet~\cite{redet} exhibits the most gradual performance degradation when subjected to any of the four types of corruptions.

\begin{figure}
    \centering
    \includegraphics[width=\linewidth]{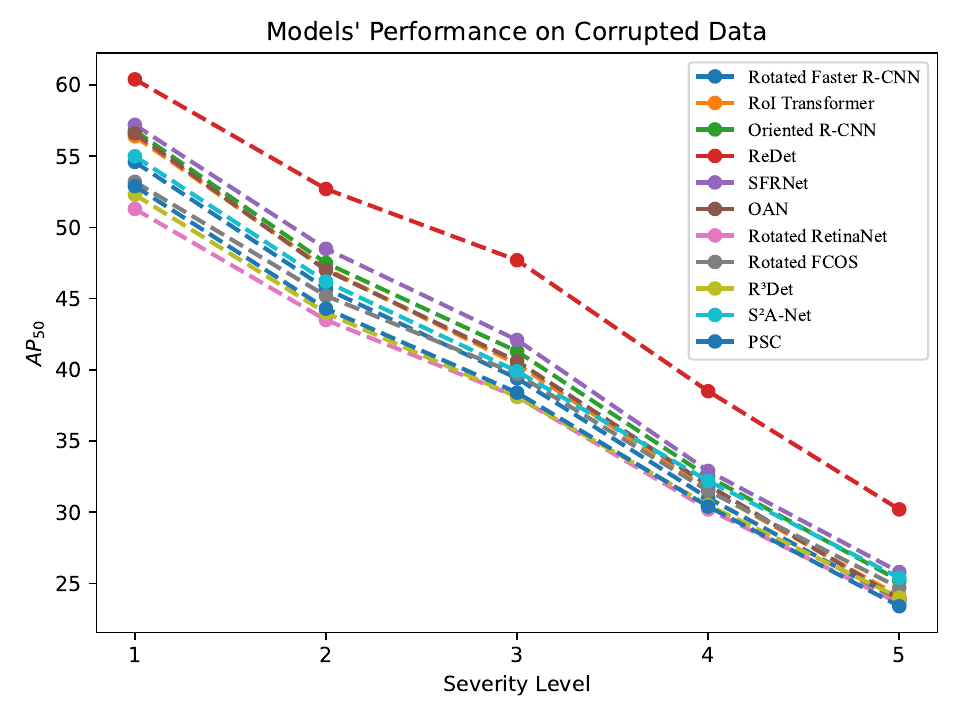}
    \caption{Evaluation of models' performance corresponding to different corruption levels on the DOTA-C dataset. The $\mathrm{AP_{50}}$ on the left is computed as the average value for all corruptions at a fixed severity level indicated by the horizontal coordinate.}
    \label{fig:models}
\end{figure}

Finally, the models' performance on corrupted data at 5 different severity levels can be seen in Figure \ref{fig:models}. With the escalation of corruption severity, a discernible reduction in models' $\mathrm{AP_{50}}$ is noted. Particularly noteworthy is the consistent manifestation of superior performance by ReDet \cite{redet}.

As for the DOTA-Cloudy dataset, the evaluation of models' performance can be seen in Table \ref{table:clouds_results}. For $\mathrm{AP}^{\text {clouds}}_{50}$, $\mathrm{S^2A}$-Net~\cite{s2anet} achieves the largest 73.9\% in the one-stage models. The two-stage models have an overall better performance, among which ReDet~\cite{redet} performs the best. For $\mathrm{rPC}_{\text {clouds}}$, ReDet~\cite{redet} is still the best and outperforms the worst model RoI Transformer~\cite{roitrans} by more than seven points. $\mathrm{R^3Det}$~\cite{r3det} has moderate detection results quantified in $\mathrm{AP}^{\text {clouds}}_{50}$, while exhibits the least performance degradation among one-stage models. 
The comparison of robustness for the tested models follows a similar pattern to that of the DOTA-C dataset.

\begin{table}[]
\begin{center}
\caption{Quantitative evaluation results of tested models on the DOTA-Cloudy dataset.}
\label{table:clouds_results}
\begin{tabular}{l|lll}
\hline
Model  & $\mathrm{AP}^{\text {clean}}_{50}$ & $\mathrm{AP}^{\text {clouds}}_{50}$   & $\mathrm{rPC}_{\text {clouds}}(\%)$      \\ \hline
Rotated Faster R-CNN \cite{fasterrcnn} & \multicolumn{1}{c}{73.4} & \multicolumn{1}{c}{58.53} & \multicolumn{1}{c}{79.73} \\ 
RoI Transformer \cite{roitrans} & \multicolumn{1}{c}{76.1} & \multicolumn{1}{c}{60.03} & \multicolumn{1}{c}{78.90} \\ 
Oriented R-CNN \cite{orientedrcnn} & \multicolumn{1}{c}{75.7} & \multicolumn{1}{c}{60.59} & \multicolumn{1}{c}{80.05} \\ 
ReDet \cite{redet}  & \multicolumn{1}{c}{76.7} & \multicolumn{1}{c}{66.19} & \multicolumn{1}{c}{86.33} \\
SFRNet \cite{sfrnet}  & \multicolumn{1}{c}{75.9}  & \multicolumn{1}{c}{60.38} & \multicolumn{1}{c}{79.55} \\ 
OAN \cite{oan}  & \multicolumn{1}{c}{73.9} & \multicolumn{1}{c}{60.25} & \multicolumn{1}{c}{81.51} \\ \hline
Rotated RetinaNet \cite{retinanet} & \multicolumn{1}{c}{68.4} & \multicolumn{1}{c}{55.12} & \multicolumn{1}{c}{80.55} \\ 
Rotated FCOS \cite{fcos} & \multicolumn{1}{c}{71.3} & \multicolumn{1}{c}{57.51} & \multicolumn{1}{c}{80.68} \\ 
$\mathrm{R^3Det}$ \cite{r3det} & \multicolumn{1}{c}{69.8} & \multicolumn{1}{c}{56.65} & \multicolumn{1}{c}{81.15} \\ 
$\mathrm{S^2A}$-Net \cite{s2anet} & \multicolumn{1}{c}{73.9} & \multicolumn{1}{c}{59.29} & \multicolumn{1}{c}{80.22} \\
PSC \cite{psc}  & \multicolumn{1}{c}{71.9} & \multicolumn{1}{c}{57.25} & \multicolumn{1}{c}{79.62} \\ \hline
\end{tabular}
\end{center}
\end{table}

\begin{figure}
    \centering
    \includegraphics[width=0.95\linewidth]{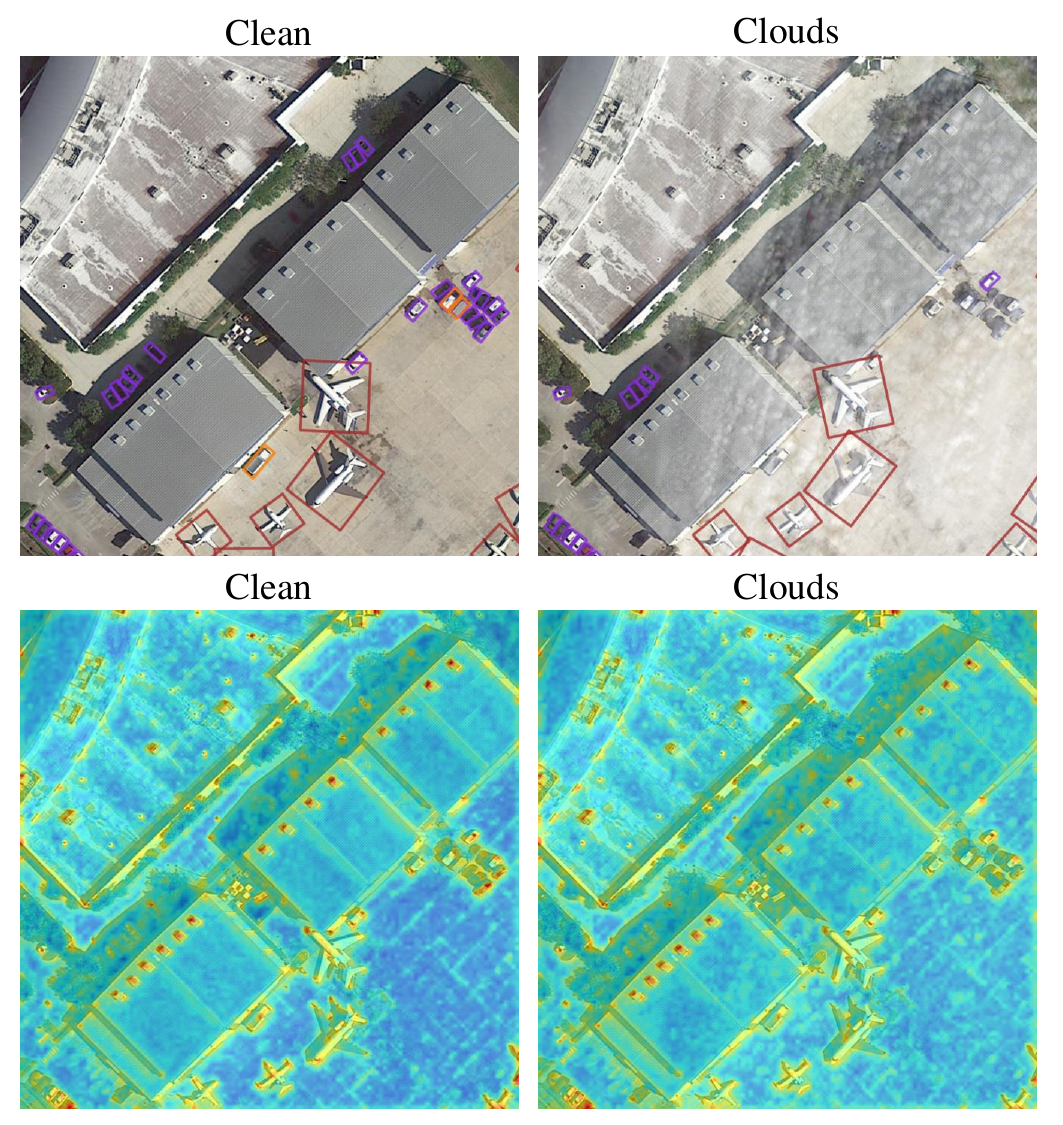}
    \caption{The detection results and heatmaps of RoI Transformer \cite{roitrans} on a cropped clean test image and the same image corrupted with clouds.}
    \label{fig:heatmap}
\end{figure}

Comparing the values of \({\rm AP_{50}^{{\rm clouds}}}\) in Table~\ref{table:clouds_results} with values of other corruption types in Table~\ref{table:mpc}, it can be seen that tested models perform relative well against the Clouds corruption. For clarity, we visualized the comparison results of RoI Transformer \cite{roitrans} in Figure \ref{fig:heatmap}, in which the heatmaps are obtained by converting some sampled feature maps from the model's backbone. Unlike other types of corruptions, clouds only affect certain areas of images. For areas not covered by clouds in the corrupted image, the detection results are almost identical to those of the clean image.
Compared to clean data, the color of objects covered by clouds in the heatmap becomes lighter, indicating that the model's confidence in the presence of objects in those areas is reduced. Meanwhile, the background color deepens where clouds are present, leading to more false positives. In addition, taking into account objects obscured by clouds in the displayed aerial image, the detection of larger objects like planes tends to be less affected than small-scale objects, such as small vehicles.

We find that on the $\mathrm{AP}_{50}^{{\rm clean}}$, $\mathrm{mPC}$, and $\mathrm{AP}^{\text {clouds}}_{50}$ metrics, two-stage models outperform one-stage models. This indicates that using RPN (Region Proposal Network) \cite{fasterrcnn} to generate high-quality region proposals can enhance the model performance. For one-stage models, $\mathrm{S^2A}$-Net \cite{s2anet} performs the best on these three metrics. Given that all one-stage models utilize ResNet50 as their backbone, we attribute the performance differences primarily to the classifier. The superior performance of $\mathrm{S^2A}$-Net's classifier may stem from its ability to align convolutional features with anchor boxes and use well-designed modules to mitigate inconsistencies between classification scores and localization accuracy.

Moreover, we can observe that ReDet \cite{redet} significantly outperforms other tested models on both DOTA-C and DOTA-Cloudy datasets as shown in Table \ref{table:mpc} and Table \ref{table:clouds_results}.
We have performed an in-depth analysis to dissect the factors contributing to this performance. It is known that ReDet~\cite{redet} is a subsequent work based on RoI Transformer~\cite{roitrans} and the core idea behind them is to extract rotation-invariant features of objects. Different from RoI Transformer~\cite{roitrans} that uses the Rotated RoI Align module to directly extract features from feature maps obtained by regular CNN, ReDet~\cite{redet} uses rotation-equivariant networks to obtain rotation-equivariant features first. Then, ReDet~\cite{redet} employs the RiRoI Align module, which can dynamically extract rotation-invariant features from rotation-equivariant features according to the orientation of the RoI. This observation suggests that using rotation-invariant modeling can enhance the robustness of object detection models.

\subsection{Ablation Study on impact factors}
We further conduct a series of ablation experiments to investigate some possible factors that may have some help on the model's robustness. RoI Transformer \cite{roitrans} is selected as the basic model because of its worst relative performance against corruption. Specifically, we analyze the effects of backbones with varying structure and capacity, and data augmentation strategies.

\subsubsection{Different Backbones}

We choose ResNet \cite{resnet}, ConvNeXt \cite{convnext}, and Swin-Transformer \cite{swin} as three kinds of backbone networks for RoI Transformer \cite{roitrans} and test their performance on the two constructed datasets. They are the typical network architecture of the CNN and Vision Transformer. For ResNet~\cite{resnet} serials, the representative models ResNet50, ResNet101, and ResNet152 are used. For ConvNeXt~\cite{convnext} serials, ConvNeXt-T is chosen to serve as a contrast to ResNet50. For Swin-Transformer~\cite{swin} serials, we adopt Swin-T, Swin-S, Swin-B, and Swin-L. The parameters of all eight models are listed in Table \ref{table:backbones}. The model size is measured by the number of parameters (\#params) and the computational complexity is represented by floating point operations per second (FLOPs) for \(224\times224\) image resolution.
We initialize the parameters of backbone networks with pre-trained weights. Apart from Swin-L, who is pre-trained on ImageNet-22K~\cite{imagenet} and then finetuned on ImageNet-1K~\cite{imagenet}, all the other backbones are pre-trained on ImageNet-1K \cite{imagenet}.  \par

\begin{table}[]
\begin{center}
\caption{Comparison of backbone network parameters.}
\label{table:backbones}
\begin{tabular}{l|lll}
\hline
Backbone & Resolution    & \#params    & FLOPs      \\ \hline
ResNet50 \cite{resnet} & \multicolumn{1}{c}{224x224} & \multicolumn{1}{c}{26M} & \multicolumn{1}{c}{3.5G} \\
ResNet101 \cite{resnet} & \multicolumn{1}{c}{224x224} & \multicolumn{1}{c}{45M} & \multicolumn{1}{c}{7.8G} \\
ResNet152 \cite{resnet} & \multicolumn{1}{c}{224x224} & \multicolumn{1}{c}{60M} & \multicolumn{1}{c}{11.5G} \\ \hline
ConvNeXt-T \cite{convnext} & \multicolumn{1}{c}{224x224} & \multicolumn{1}{c}{28M} & \multicolumn{1}{c}{4.5G} \\ \hline
Swin-T \cite{swin} & \multicolumn{1}{c}{224x224} & \multicolumn{1}{c}{28M} & \multicolumn{1}{c}{4.5G} \\
Swin-S \cite{swin} & \multicolumn{1}{c}{224x224} & \multicolumn{1}{c}{50M} & \multicolumn{1}{c}{8.7G} \\
Swin-B \cite{swin} & \multicolumn{1}{c}{224x224} & \multicolumn{1}{c}{88M} & \multicolumn{1}{c}{15.4G} \\ 
Swin-L \cite{swin} & \multicolumn{1}{c}{224x224} & \multicolumn{1}{c}{197M} & \multicolumn{1}{c}{34.5G} \\ \hline
\end{tabular}
\end{center}
\end{table}

\begin{table}[]
\begin{center}
\caption{Ablation study on backbones on the two robustness benchmark datasets.}
\label{table:backbones_results}
\resizebox{\linewidth}{!}{
\begin{tabular}{l|llllll}
\hline
Backbone & $\mathrm{AP}^{\text {clean}}_{50}$   & $\mathrm{mPC}$   & $\mathrm{rPC}$(\%)     & $\mathrm{AP}^{\text {clouds}}_{50}$   & $\mathrm{rPC}_{\text {clouds}}(\%)$      \\ \hline
ResNet50 \cite{resnet} & \multicolumn{1}{c}{76.08} & \multicolumn{1}{c}{39.92} & \multicolumn{1}{c}{52.46}  & \multicolumn{1}{c}{60.03} & \multicolumn{1}{c}{78.90} \\ 
ResNet101 \cite{resnet} & \multicolumn{1}{c}{74.34}  & \multicolumn{1}{c}{38.80} & \multicolumn{1}{c}{52.18}  & \multicolumn{1}{c}{57.96} & \multicolumn{1}{c}{77.97} \\
ResNet152 \cite{resnet} & \multicolumn{1}{c}{75.11}  & \multicolumn{1}{c}{39.48} & \multicolumn{1}{c}{52.56}  & \multicolumn{1}{c}{59.74} & \multicolumn{1}{c}{79.54} \\ \hline
ConvNeXt-T \cite{convnext} & \multicolumn{1}{c}{74.98} & \multicolumn{1}{c}{47.81} & \multicolumn{1}{c}{63.76}  & \multicolumn{1}{c}{64.52} & \multicolumn{1}{c}{86.04} \\ \hline
Swin-T \cite{swin} & \multicolumn{1}{c}{77.51} & \multicolumn{1}{c}{43.38} & \multicolumn{1}{c}{55.97}  & \multicolumn{1}{c}{62.83} & \multicolumn{1}{c}{81.06} \\
Swin-S \cite{swin} & \multicolumn{1}{c}{77.12} & \multicolumn{1}{c}{44.54} & \multicolumn{1}{c}{57.75}  & \multicolumn{1}{c}{63.27} & \multicolumn{1}{c}{82.05} \\
Swin-B \cite{swin} & \multicolumn{1}{c}{77.70} & \multicolumn{1}{c}{45.06} & \multicolumn{1}{c}{57.99}  & \multicolumn{1}{c}{64.84} & \multicolumn{1}{c}{83.45} \\
Swin-L \cite{swin} & \multicolumn{1}{c}{77.65} & \multicolumn{1}{c}{47.76} & \multicolumn{1}{c}{61.50}  & \multicolumn{1}{c}{66.73} & \multicolumn{1}{c}{85.94} \\ \hline
\end{tabular}}
\end{center}
\end{table}

The evaluation results of RoI Transformer~\cite{roitrans} with different backbones are demonstrated in Table \ref{table:backbones_results}. First, we select ResNet50~\cite{resnet}, ConvNeXt-T~\cite{convnext}, and Swin-T~\cite{swin} for comparison for similar model complexity.
Among the three models, ConvNeXt-T~\cite{convnext} has the lowest value of $\mathrm{AP}^{\text {clean}}_{50}$ and Swin-T~\cite{swin} achieves the largest. Despite the failure in improving object detection results on clean data, the architectural change from ResNet~\cite{resnet} to ConvNeXt~\cite{convnext} and Swin Transformer~\cite{swin} helps to enhance the model's robustness. ConvNeXt-T~\cite{convnext} outperforms the others in terms of \({\rm mPC}\) and \({\rm AP}_{50}^{\rm clouds}\) metrics for both corrupted datasets. Notably, the highest \({\rm rPC}\) and \({\rm rPC}_{\rm clouds}\) indicate that ConvNeXt-T~\cite{convnext} also has a minimum of performance degradation in the presence of corruption. Besides, Table \ref{table:extend} reveals that ConvNeXt-T performs much better than the other two kinds of networks for all corruption types especially Gaussian, Shot, Impulse, and Speckle noise. Its performance even matches the largest model Swin-L~\cite{swin}. Generally, it is justifiable to hypothesize that the architecture of ConvNeXt~\cite{convnext} exhibits higher stability and resilience against corruption. \par

\begin{table*}[]
\begin{center}
\caption{Comparison of detailed results of RoI Transformer~\cite{roitrans} with various backbones for every type of corruption in the DOTA-C dataset.}
\vspace{-10pt}
\label{table:extend}
\resizebox{\linewidth}{!}{
\begin{tabular}{l|ccccccccccccccccccc}
\hline
 \multirow{2}*{Backbone}          & \multicolumn{1}{l}{}  & \multicolumn{2}{c}{Noise} & \multicolumn{1}{l}{}     & \multicolumn{1}{l}{} & \multicolumn{1}{l}{} & Blur & \multicolumn{1}{l}{} & \multicolumn{1}{l}{}      & \multicolumn{1}{l}{} & \multicolumn{1}{l}{} & Weather & \multicolumn{1}{l}{} & \multicolumn{1}{l}{}      & \multicolumn{1}{l}{} & \multicolumn{1}{l}{} & Digital & \multicolumn{1}{l}{} & \multicolumn{1}{l}{} \\ \cline{2-20}
  & Ga. & Shot & Im.   & \multicolumn{1}{c|}{Spec.} & De.  & Glass  & Mo.  & Zoom   & \multicolumn{1}{c|}{Ga.} & Snow  & Frost  & Fog  & Br.   & \multicolumn{1}{c|}{Spat.} & Co.  & El.  & Pixel  & JPEG  & Sa.         \\ \hline
ResNet50 \cite{resnet}   & 19.8  & 20.2 & 17.8  & \multicolumn{1}{c|}{29.1}   & 41.1  & 48.8  & 42.6   & 14.7   & \multicolumn{1}{c|}{44.0}   & 26.5  & 47.1  & 49.2   & 63.5   & \multicolumn{1}{c|}{49.4}   & 42.5  & 35.0  & 53.6   & 51.5   & 62.3 \\ 
ResNet101 \cite{resnet}   & 19.1  & 19.2 & 15.5  & \multicolumn{1}{c|}{26.9}   & 42.0  & 47.6  & 42.1   & 13.5   & \multicolumn{1}{c|}{45.2}   & 25.4  & 45.1  & 48.4   & 62.1   & \multicolumn{1}{c|}{46.3}   & 40.8  & 34.9  & 51.8   & 50.0   & 61.3 \\
ResNet152 \cite{resnet}   & 20.7  & 21.2 & 17.5  & \multicolumn{1}{c|}{29.1}   & 41.7  & 46.9  & 41.3   & 13.2   & \multicolumn{1}{c|}{44.6}   & 28.1  & 48.4  & 49.6   & 63.8   & \multicolumn{1}{c|}{46.8}   & 42.7  & 33.6  & 50.6   & 48.2   & 62.1 \\ \hline
ConvNeXt-T \cite{convnext}    & 33.3  & 33.1 & 31.9  & \multicolumn{1}{c|}{43.2}   & 47.8  & 52.0  & 49.1   & 17.7   & \multicolumn{1}{c|}{50.7}   & 37.0  & 56.3  & 63.5   & 68.9   & \multicolumn{1}{c|}{54.8}   & 56.0  & 32.0  & 56.7   & 58.5   & 65.8 \\ \hline
Swin-T \cite{swin}    & 26.3  & 25.8  & 25.9  & \multicolumn{1}{c|}{35.9}   & 45.6  & 48.7  & 44.0   & 15.6   & \multicolumn{1}{c|}{48.4}   & 34.9  & 53.4  & 59.4   & 67.4   & \multicolumn{1}{c|}{55.1}   & 48.3  & 37.4  & 53.9   & 55.7   & 62.4 \\
Swin-S \cite{swin}  & 26.3  & 25.8  & 25.9  & \multicolumn{1}{c|}{35.9}   & 45.6  & 48.7  & 46.1   & 15.6   & \multicolumn{1}{c|}{48.4}   & 34.9  & 53.4  & 59.4   & 67.4   & \multicolumn{1}{c|}{55.1}   & 48.3  & 37.4  & 53.9   & 55.7   & 62.4 \\
Swin-B \cite{swin}   & 26.6  & 26.6  & 27.4  & \multicolumn{1}{c|}{36.4}   & 47.1  & 50.7  & 46.2   & 15.0   & \multicolumn{1}{c|}{49.7}   & 35.5  & 54.5  & 58.0   & 68.7   & \multicolumn{1}{c|}{55.2}   & 48.5  & 37.6  & 50.6   & 57.4   & 64.5 \\ 
Swin-L \cite{swin}   & 32.5  & 32.1  & 33.9  & \multicolumn{1}{c|}{42.1}   & 46.8  & 50.1  & 46.7  & 16.4   & \multicolumn{1}{c|}{49.9}   & 35.9  & 58.3  & 63.8   & 70.6   & \multicolumn{1}{c|}{57.7}   & 53.1  & 37.1  & 55.3   & 58.6   & 66.5 \\ \hline
\end{tabular}}
\end{center}
\end{table*}

\begin{figure}[]
    \centering
    \includegraphics[width=0.95\linewidth]{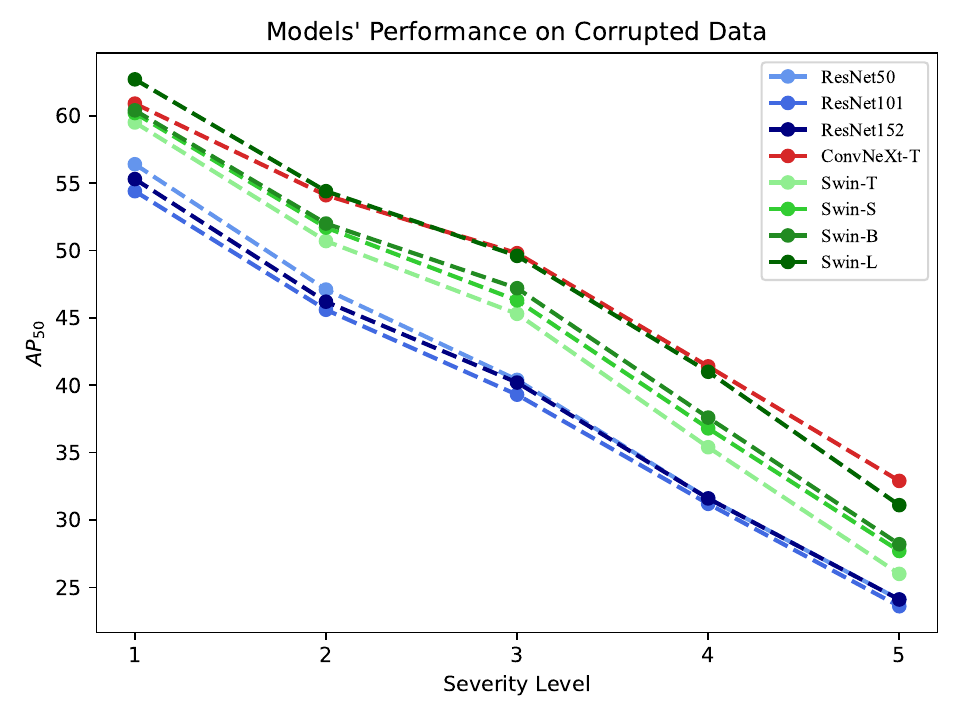}
    \caption{Evaluation results of RoI Transformer~\cite{roitrans} with different backbones at five severity levels of corruption in the DOTA-C dataset.}
    \label{fig:roi_trans}
\end{figure}

On the basis of CNN-based network ResNet \cite{resnet} and Transformer-based network Swin-Transformer \cite{swin}, we analyze their results respectively to investigate the impact of different backbone capacities on the model's performance. For ResNet serials~\cite{resnet} in Table~\ref{table:backbones_results}, the performance metrics of ResNet101~\cite{resnet} are inferior to those of ResNet50~\cite{resnet}. ResNet152~\cite{resnet} experiences a decrease in $\mathrm{AP}^{\text {clean}}_{50}$, \({\rm mPC}\), and $\mathrm{AP}^{\text {clouds}}_{50}$, while showing an improvement in $\mathrm{rPC}$ and $\mathrm{rPC}_{\text {clouds}}$. Swin-Transformer~\cite{swin} benefits from its scalability \cite{vit}. From Swin-T~\cite{swin} to Swin-L~\cite{swin}, their results on corrupted data gradually improve across various metrics as the model becomes deepened and expanded. The trend of performance change is more evident in Figure~\ref{fig:roi_trans}, which demonstrates the object detection results of RoI Transformer~\cite{roitrans} with different backbones under different levels of corruption.
The results indicate that incorporating a larger and deeper Transformer-based backbone into the model contributes positively to enhancing its robustness, under the assumption of keeping other influencing factors unchanged. 

\subsubsection{Data Augmentations} 
\begin{figure}[h]
    \centering
    \includegraphics[width=\linewidth]{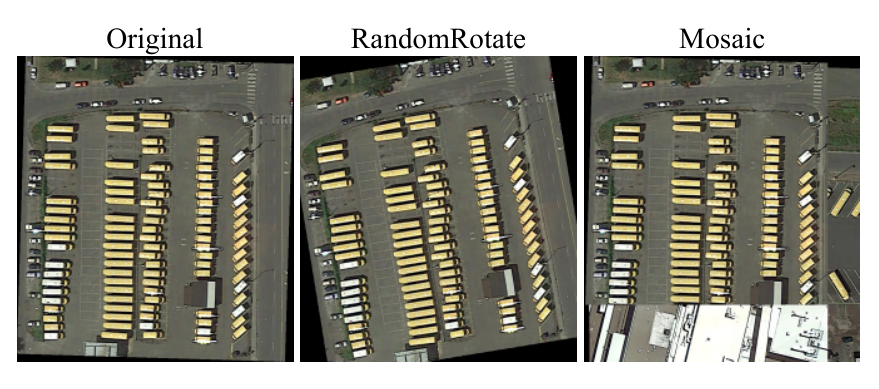}
    \caption{From left to right, it is the original image, the image after using RandomRotate and the image after using Mosaic.}
    \label{fig:augmentation}
\end{figure}

We aim to explore the effects of general data augmentations unrelated to these visual corruptions on models' robustness. In order to ensure that these corruptions are unseen to the models during training, we choose two data augmentation strategies, RandomRotate and Mosaic \cite{yolov4} as shown in Figure \ref{fig:augmentation}. RandomRotate means to rotate the images randomly. And Mosaic is defined as combining multiple images into a cohesive mosaic. During training, four aerial images are combined in a certain scale to form a single image. 
To improve the generality and universality of the experiments, we additionally select a typical one-stage object detection model, Rotated RetinaNet~\cite{retinanet}, for evaluation. It is worth mentioning that RoI Transformer~\cite{roitrans} and Rotated RetinaNet~\cite{retinanet} both use the backbone ResNet50~\cite{resnet}, and their training and testing settings are the same.

The comparison results of data augmentation techniques are displayed in Table~\ref{table:augmentation}. RoI Transformer \cite{roitrans} benefits from the application of RandomRotate in model training, which leads to a comprehensive enhancement in all performance evaluation metrics. However, there is no similar improvement for the one-stage model Rotated RetinaNet~\cite{retinanet} as measured by the reduced values of \({\rm mPC}\) and $\mathrm{AP}^{\text {clouds}}_{50}$. Beyond that, the way of creating new data by using Mosaic is extremely limited. The addition of Mosaic brings a slightly bigger score of \({\rm rPC}\) and \({\rm rPC_{\text{clouds}}}\) for RoI Transformer~\cite{roitrans}, and only \({\rm rPC_{\text{clouds}}}\) for Rotated RetinaNet~\cite{retinanet}. Despite these, using Mosaic even decreases the \({\rm mPC}\) scores of both models. 
In contrast with Mosaic, RandomRotate tends to be a better choice.

\begin{table*}[]
\begin{center}
\caption{Ablation study on data augmentations on the two robustness benchmark datasets.}
\label{table:augmentation}
\begin{tabular}{l|clllll}
\hline
Model & Augmentation & $\mathrm{AP}^{\text {clean}}_{50}$   & $\mathrm{mPC}$   & rPC(\%)    & $\mathrm{AP}^{\text {clouds}}_{50}$   & $\mathrm{rPC}_{\text {clouds}}(\%)$      \\ \hline
\multirow{3}*{RoI Transformer~\cite{roitrans}} & - &\multicolumn{1}{c}{76.08} & \multicolumn{1}{c}{39.92} & \multicolumn{1}{c}{52.46}  & \multicolumn{1}{c}{60.03} & \multicolumn{1}{c}{78.90}\\
& RandomRotate & \multicolumn{1}{c}{76.38} & \multicolumn{1}{c}{41.22} & \multicolumn{1}{c}{53.96}  & \multicolumn{1}{c}{61.26} & \multicolumn{1}{c}{80.21}\\
& Mosaic & \multicolumn{1}{c}{74.36} & \multicolumn{1}{c}{39.08} & \multicolumn{1}{c}{52.56}  & \multicolumn{1}{c}{59.56} & \multicolumn{1}{c}{80.10} \\ \hline
\multirow{3}*{Rotated RetinaNet \cite{retinanet}} &- & \multicolumn{1}{c}{68.43}  & \multicolumn{1}{c}{37.34} & \multicolumn{1}{c}{54.57}  & \multicolumn{1}{c}{55.12} & \multicolumn{1}{c}{80.55} \\
&RandomRotate & \multicolumn{1}{c}{66.44} & \multicolumn{1}{c}{36.48} & \multicolumn{1}{c}{54.91}  & \multicolumn{1}{c}{53.71} & \multicolumn{1}{c}{80.84} \\
&Mosaic & \multicolumn{1}{c}{67.60} & \multicolumn{1}{c}{36.01} & \multicolumn{1}{c}{53.27}  & \multicolumn{1}{c}{54.47} & \multicolumn{1}{c}{80.58} \\ \hline
\end{tabular}
\end{center}
\end{table*}

\begin{figure*}[]
    \centering
    \includegraphics[width=0.9\textwidth]{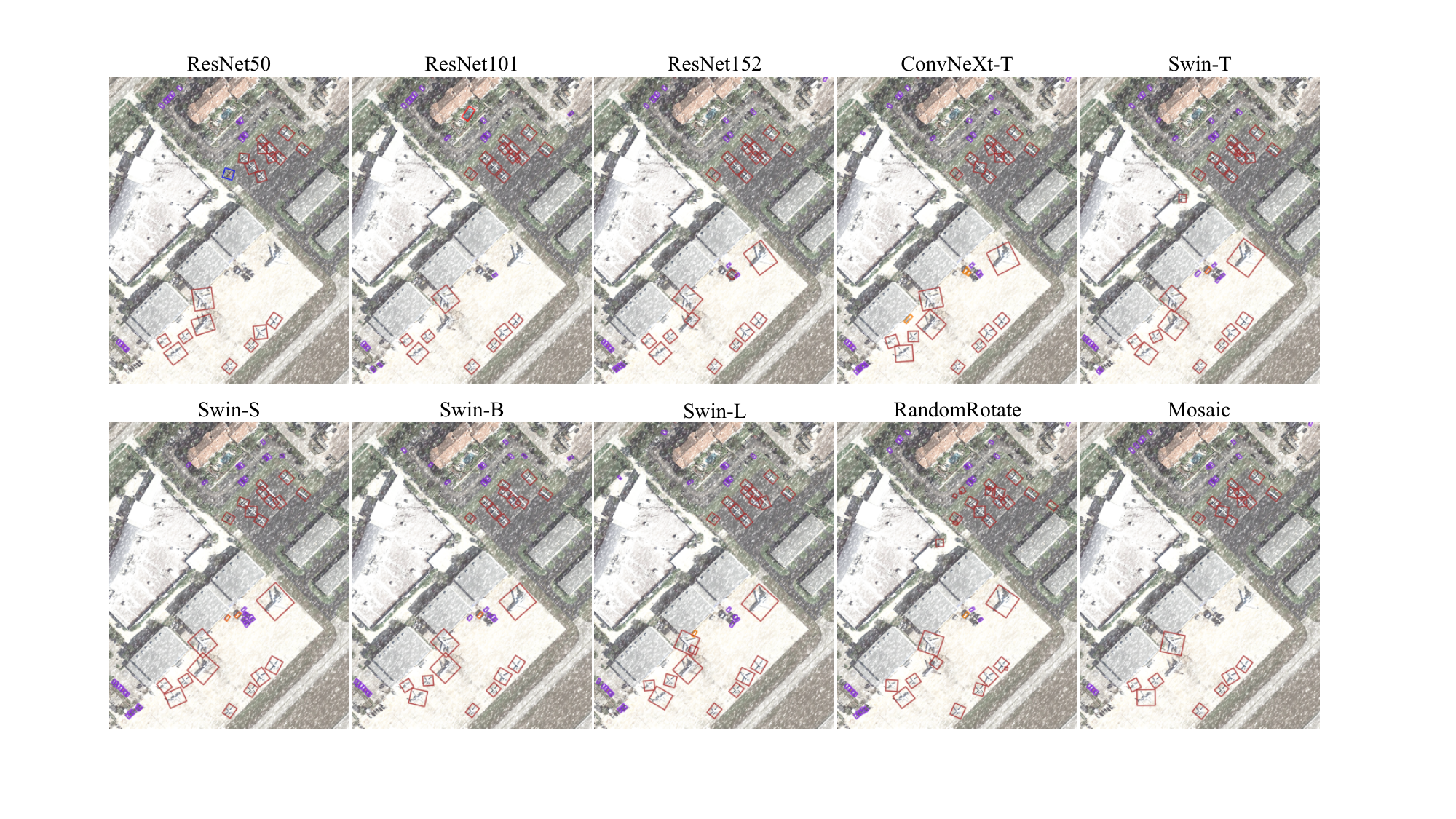}
    \caption{Comparison of object detection results of RoI Transformer~\cite{roitrans} considering variances in backbone networks and adding data augmentation techniques for a sampled image corrupted with Snow at severity level 3. The first image in the upper left corner annotated as ResNet50 refers to RoI Transformer~\cite{roitrans} using ResNet50~\cite{resnet} as the backbone with plain training strategy, which serves as the basic. These images only display bounding boxes with a confidence score of no less than 0.3.}
    \vspace{-10pt}
    \label{fig:ablation}
\end{figure*}

In general, the effectiveness of the methods mentioned above in improving the performance of RoI Transformer \cite{roitrans} on corrupted data is not guaranteed and may vary. The visualized object detection results are shown in Figure~\ref{fig:ablation}. We can see that replacing the backbone with ConvNeXt-T~\cite{convnext} or Swin-L~\cite{swin} has the most significant improvement, whereas the utilization of data augmentation exhibits limited positive impact. 

\section{Conclusion}

In this paper, we propose two new benchmarks for investigating the robustness of models on corrupted data in aerial object detection. We apply the 19 common corruption types to form the DOTA-C dataset and simulate cloud cover to generate the DOTA-Cloudy dataset.
Based on the two datasets, we conduct a detailed evaluation of numerous mainstream models' robustness and find that the models suffer varying degrees of performance degradation for unknown image corruptions. Models that perform better on clean data do not necessarily perform better on corrupted data. Yet rotation-invariant modeling proves to be pivotal for achieving accurate and robust object detection. Then, a series of ablation experiments are carried out to demonstrate that some modifications can effectively improve the robustness of models, such as enhanced backbone architectures. For Transformer-based backbones, increasing the capacity can strengthen their robustness. 

Future research should give more consideration to the specific types of corruptions in aerial imagery, such as geometric distortions of objects due to changes in viewpoint, and shadows on the ground created by variations in solar elevation and azimuth angles, etc. These corruptions commonly appear in aerial images and can reduce image quality to varying degrees. In this way, more realistic, comprehensive, and challenging benchmark datasets can be created for evaluating and improving the models' performance. We hope that our benchmarks and study can provide some help for future research on the robustness of models in aerial object detection.

\bibliographystyle{IEEEtran}
\bibliography{IEEEabrv,dotac}

\end{document}